\documentclass{article}

\usepackage{arxiv}

\usepackage[utf8]{inputenc} % allow utf-8 input
\usepackage[T1]{fontenc}    % use 8-bit T1 fonts
\usepackage{hyperref}       % hyperlinks
\usepackage{url}            % simple URL typesetting
\usepackage{booktabs}       % professional-quality tables
\usepackage{amsmath,amsfonts,amssymb}
\usepackage{nicefrac}       % compact symbols for 1/2, etc.
\usepackage{microtype}      % microtypography
\usepackage{lipsum}
\usepackage{graphicx}
\graphicspath{ {./images/} }
\usepackage{bbm}
\usepackage{array}
\usepackage{hyperref}
\usepackage{color}
\usepackage{stfloats}
\usepackage{algorithm}
\usepackage{algpseudocode}

\usepackage{xcolor}

\usepackage{colortbl}
\newcolumntype{P}[1]{>{\centering\arraybackslash}p{#1}}
\newcolumntype{R}[1]{>{\flushrifht\arraybackslash}p{#1}}

\definecolor{c_diag}{rgb}{0.85,0.89,0.953}
\definecolor{t_diag}{rgb}{0,0,0}
\definecolor{c_ndiag}{rgb}{.95,0.95,0.95}

\definecolor{c_hlight}{rgb}{.886,0.9411,0.851}
\definecolor{c_hdark}{rgb}{.663,0.82,0.557}
\definecolor{c_rest}{rgb}{.984,0.898,0.839}

\title{Irredundant  k--Fold Cross--Validation}

\author{
 Jes\'us S. Aguilar--Ruiz \\
  School of Engineering\\ Pablo de Olavide University\\ 
  ES-41013 Seville, Spain \\
  \texttt{aguilar@upo.es} \\
 }

\begin{document}
\maketitle
\begin{abstract}
In traditional k--fold cross--validation, each instance is used ($k\!-\!1$) times for training and once for testing, leading to redundancy that lets many instances disproportionately influence the learning phase. We introduce Irredundant $k$--fold cross--validation, a novel method that guarantees each instance is used exactly once for training and once for testing across the entire validation procedure. This approach ensures a more balanced utilization of the dataset, mitigates overfitting due to instance repetition, and enables sharper distinctions in comparative model analysis. The method preserves stratification and remains model--agnostic, i.e., compatible with any classifier. Experimental results demonstrate that it delivers consistent performance estimates across diverse datasets --comparable to $k$--fold cross--validation-- while providing less optimistic variance estimates because training partitions are non--overlapping, and significantly reducing the overall computational cost.
\end{abstract}

% keywords can be removed

\section{Introduction}

Validation is a cornerstone in the development of predictive models in machine learning. As algorithms are increasingly deployed in high--stakes applications---from medical diagnosis to financial forecasting---the need for rigorous evaluation methodologies has become critical. The primary aim of validation is to estimate the generalization capability of a trained model, i.e., its performance on unseen data. Without proper validation, performance metrics derived from training data may be overly optimistic because of overfitting or unnecessarily pessimistic because of underfitting \cite{Langley1988}.

A good error estimator exhibits both small bias (i.e., a small expected deviation of the estimated performance from the true performance) and small variance (i.e., low sensitivity of the estimate to the specific partitioning of the data). 

%\subsubsection*{Resubstitution}

The earliest and simplest validation approach is the \emph{resubstitution} method \cite{Larson1931,Smith1947}, which uses the same dataset both for training and evaluation. However, the resubstitution estimator is usually optimistically biased: the difference between the estimate and the true generalization error, termed the \emph{generalization gap} \cite{Keskar2017}, is negative with high probability. Consequently, it is highly biased as an estimator of the true error \cite{Devroye1996}. Complex classifiers, in particular, tend to overfit the data, especially with small samples, leading to highly optimistic resubstitution errors \cite{Vapnik1999}.

%\subsubsection*{Hold--out}

The \emph{hold--out} method \cite{Devroye1979} improves on this by partitioning the data into two disjoint subsets: one for training and another for testing (typically, $2/3$ for training and $1/3$ for testing). The hold--out error estimator is unbiased for the generalization error provided that the test data are truly independent of the training data and have not been reused in any form \cite{BragaNeto2015, Yousefi2011}. While computationally inexpensive, the hold--out method is sensitive to how the data are split, resulting in high variance in the estimated metrics. 

%\subsubsection*{k--Fold cross--validation}

To address the instability of hold--out, \emph{$k$--fold cross--validation} was proposed \cite{Geisser1975}. In this method, the data is partitioned into $k$ disjoint subsets (folds). Each fold serves once as a test set, while the remaining $k-1$ folds are used for training. The final performance estimate is the average across all $k$ iterations. Cross--validation reduces variance and produces more robust estimates, particularly in small datasets. However, its variance is significantly greater than that of resubstitution \cite{BragaNeto2004c}. As noted by Nadeau \& Bengio \cite{Nadeau2003}, the use of cross--validation estimators for model selection sparked a debate \cite{Goutte1997,Zhu1996} related to the ``no free lunch theorem'' \cite{Wolpert1995}, highlighting that while cross--validation often works well in practice, it is not guaranteed to perform uniformly well across all learning problems.

An important extension of $k$--fold cross--validation is \emph{nested cross--validation}, in which an inner cross--validation loop (usually $(k\!-\!1)$--fold) is executed within each outer test fold, and the risk estimate is adjusted using the outer--fold loss \cite{Bates2024}. This recent approach provides a more reliable estimate of generalization error by accounting for the additional variability introduced by model selection or hyperparameter tuning.

%\subsubsection*{Leave--one--out}

A limiting case of $k$--fold cross--validation is \emph{leave--one--out cross--validation} \cite{Allen1974,Efron1972,Stone1974}, where each individual instance serves once as the test instance while the rest are used for training. Although this method yields minimal bias, it can be computationally intensive and prone to high variance, particularly for models with high complexity.

A generalization of leave--one--out is \emph{leave--$p$--out cross--validation} \cite{Shao1993}, where $p$ instances are left out at each iteration, and every possible subset of $p$ instances is used as a validation set. Notably, leave--one--out corresponds to leave--$p$--out with $p\!=\!1$. 

%\subsubsection*{Bootstrap}

\emph{Bootstrap methods} \cite{Efron1979} introduced a probabilistic resampling framework that enables alternative estimates of model performance. However, bootstrap--based estimators can exhibit increased bias, especially in small--sample settings.

%\subsubsection*{Bolstering}

Alternative approaches, such as \emph{bolstering}~\cite{BragaNeto2004}, enrich the empirical data distribution by applying bolstering kernels to the training set. These methods can achieve error estimators with lower variance and typically smaller bias compared to classic resampling methods.

%\subsubsection*{Imbalance: Stratified}

On the other hand, to mitigate the challenges posed by class imbalance in classification tasks, \emph{stratified} cross--validation was introduced, where each fold maintains approximately the same class distribution as the full dataset  \cite{Kohavi1995}. Stratification remains a standard practice in classification evaluation today.

%\subsubsection*{Summary}

Each validation strategy entails trade--offs in terms of computational cost, variance, and bias. Hold--out is fast but unstable; leave--one--out is unbiased but often computationally prohibitive; $k$--fold cross--validation balances variance and bias reasonably well but can still be expensive; and bootstrap methods introduce an additional stochastic layer, sometimes at the cost of increased bias, that may not always reflect real--world generalization.

As models increase in complexity and datasets become larger and more heterogeneous, more sophisticated and computationally efficient validation schemes are needed. This paper introduces \emph{irredundant $k$--fold cross--validation}, a novel approach that aims to combine statistical rigor with computational efficiency. Unlike standard $k$--fold cross--validation, where each sample is used $k\!-\!1$ times for training, irredundant $k$--fold ensures that each sample is used exactly once for training and exactly once for testing. Experimental results demonstrate that irredundant $k$--fold achieves significant computational savings while preserving the quality of performance estimates.

The evolution of validation methodologies in machine learning reflects not only increasing mathematical sophistication but also heightened awareness of practical constraints and ethical considerations in real--world deployment. Validation now plays a pivotal role beyond model evaluation---it is critical for hyper--parameter optimization, feature selection, and uncertainty quantification. Future directions point towards integrated frameworks that couple validation with explainability and fairness, positioning validation as a central pillar in trustworthy Artificial Intelligence.

%\subsubsection*{Paper Organization}

The remainder of this paper is organized as follows: Section \ref{sec:method} describes the \emph{irredundant $k$--fold cross--validation} in detail and includes, in Section \ref{sec:analysis}, a brief analysis of its bias and variance; Section \ref{sec:experiments} presents empirical results across several dataset from the UCI repository. Finally, \ref{sec:conclusions} summarizes the main conclusions and outlines future work. 

\section{Method}\label{sec:method}

\subsection{Overview}

Let $\mathcal{D} = \{(\mathbf{x}_i, y_i)\}_{i=1}^n$ denote a labeled dataset with $n$ instances. Let $k$ be the number of folds used for cross--validation. Let $\mathcal{M}$ be the predictive model, $\delta$ the performance metric (e.g., accuracy or F--score), and $\theta$ the scalar performance value returned by applying $\delta$ to model $\mathcal{M}$ on dataset $\mathcal{D}$. Denote by  $\mathcal{E}^i$ and $\mathcal{T}^i$ the training and test sets, respectively, in the $i$-th fold ($i\in\{1,\dots,k\}$).

The fundamental difference between standard $k$--fold cross--validation ($\mathsf{kF}$) and the proposed \emph{irredundant $k$--fold cross--validation} ($\mathsf{IkF}$) lies how many times each instance is used for training. In traditional $\mathsf{kF}$, each instance appears in the training set ($k-1$) out of the $k$ iterations. Therefore, per iteration:

\begin{equation}
|\mathcal{E}^i_{\mathsf{kF}}| \approx \frac{n}{k} \left(k-1\right) \qquad |\mathcal{T}^i_{\mathsf{kF}}| \approx \frac{n}{k}
\end{equation}
In contrast, for $\mathsf{IkF}$, each instance is used exactly once for training and once for testing (uniform exposure). As such, each training set consists of:
\begin{equation}
|\mathcal{E}^i_{\mathsf{IkF}}|  \approx \frac{n}{k} \qquad  |\mathcal{T}^i_{\mathsf{IkF}}| \approx \frac{n}{k}
\end{equation}

To achieve this, $\mathsf{IkF}$ introduces a subfolding strategy. Each outer fold is divided into $k-1$ subfolds. For each test fold $\mathcal{F}^{i}$, the training set $\mathcal{E}^{i}$ is constructed by selecting one unique, unused subfold from each fold $\mathcal{F}_j$ with $j \neq i$. The subfolds may be selected randomly or via heuristic criteria, as long as each is used exactly once across all folds.

Consequently, the training and test sets across iterations are disjoint, and the joining of all the training set is equal to the original dataset, as well as the joining of all the test sets:
\begin{equation}
\mathcal{E}^{i} \cap \mathcal{E}^{j} = \emptyset \qquad  \mathcal{T}^{i} \cap \mathcal{T}^{j} = \emptyset   \qquad  \text{for all } i \neq j
\end{equation}
\begin{equation}
\mathcal{D}  = \bigcup_{i=1}^k \mathcal{E}^{i} =  \bigcup_{i=1}^k \mathcal{T}^{i} 
\end{equation}

The proposed $\mathsf{IkF}$ guarantees that each instance appears exactly once in a test set $\mathcal{T}^i$ and exactly once in a training set $\mathcal{E}^i$. Furthermore, each fold $\mathcal{F}_i$ is used exactly once for testing, and each training set $\mathcal{E}^{i}$ is composed of $k\!-\!1$ distinct subfolds, one from each of the other folds. Thus, the size of each training set remains approximately $n/k$.

\subsection{Algorithm Description}

The full $\mathsf{IkF}$ procedure is summarized in Algorithm \ref{alg:ikf}. Structurally, it resembles classical $\mathsf{kF}$, but differs in Line \#7, where the training set $\mathcal{E}^i$ is built by assembling one unused subfold from each of the $k-1$ non--test folds.

Each subfold $\mathcal{F}_j^{\alpha_{ij}}$ is selected so that no subfold is reused across iterations. The value of each $\alpha_{ij}$ is unique within fold $j$, i.e., the pair $(j,\alpha_{ij})$ is unique. Therefore, each subfold $\mathcal{F}_j^{\alpha_{ij}}$ is chosen only once. 

Selection may be random or based on specific heuristics (e.g., forward sequential selection or optimization over diversity or stratification). When stratified splitting is used in Lines \#1 and \#3, the class distributions within folds and subfolds mirror that of the original dataset. 

\begin{algorithm}[tb]
\caption{Irredundant \textit{k}-Fold Cross--Validation ($\mathsf{IkF}$).}
\label{alg:ikf}
\textbf{Input:} $\mathcal{D}$: Data; $k$: number of folds; $\mathcal{M}$: predictive model; $\delta$: metric\\
\textbf{Output:} $\theta$: Performance metric value averaged over $k$ folds. 
\begin{algorithmic}[1]
\State Split $\mathcal{D}$ into $k$ folds: $\{\mathcal{F}_1, \dots, \mathcal{F}_k\}$.
\For{$i = 1$ to $k$}
    \State Split $\mathcal{F}_i$ into $k$$-$$1$ subfolds $\mathcal{F}_i^{j}$: $\mathcal{F}_i = \bigcup_{j=1}^{k-1} \mathcal{F}_i^{j}$
\EndFor
\For{$i = 1$ to $k$}
    \State $\mathcal{T}^{i}\! \gets\! \mathcal{F}_i$ \Comment{Test set}
    \State $\mathcal{E}^{i}\! \gets\! \bigcup_{j\neq i} \mathcal{F}_j^{\alpha_{ij}}$ where $\alpha_{ij}\!\in\!\{1,\!\dots\!,k\!-\!1\}$ \Comment{Training set} 
%    \State $\mathcal{E}^{i} \gets \emptyset$ \Comment{Training set}
%    \For{$j = 1$ to $k$}
%        \If{$j \neq i$}
%            \If{$j > i$}
%               \State $w = i$
%            \Else
%               \State $w = i-1$	    
%            \EndIf
 %           \State Random selection of $\mathcal{F}_j^{w}$$\in$$\mathcal{F}_j$ not yet used 
 %           \State $\mathcal{E}^{i} \gets \mathcal{E}^{i} \cup \mathcal{F}_j^{w}$ \Comment{Training set}
 %       \EndIf
 %   \EndFor
    \State Train model $\mathcal{M}^{i}$ on $\mathcal{E}^{i}$
    \State $\theta^{i}$ $\gets$ Evaluate $\mathcal{M}^{i}$ on $\mathcal{T}^{i}$ with $\delta$ \Comment{Performance} 
\EndFor
\State $\theta  \gets \frac{1}{k} \sum_{i=1}^{k} \theta ^{i}$ \Comment{Compute final performance}
\end{algorithmic}
\end{algorithm}

%Because each fold contains $k\!-\!1$ subfolds, the number of instances at each subfold is approximately $n/[k(k-1)]$, so the irredundant scheme requires $n \geq k(k-1)$. For stratified $\mathsf{IkF}$, a stricter constraint holds: $n \geq k(k-1)c$, where $c$ is the number of classes; otherwise, it cannot be guaranteed that each subfold contains at least one instance from each class. in addition, in case of imbalance, the class with less cardinality ($c_{min}$; minority class) should have enough instances to be present in each fold, i.e., $c_{min}\geq k(k-1)$. For instance, for $k=10$ and $c=3$, a minimum of $n=270$ instances is required, and the minority class should have at least 90 instances. These constraints for stratified $\mathsf{IkF}$ are mild for most real--world datasets. Only in cases of great imbalance, the value of $k$ should be reduced (e.g., $k=5$).

Because each fold is split into $k\!-\!1$ subfolds, the expected size of a subfold is $n/[k(k-1)]$. Hence the irredundant scheme is feasible only if $n \geq k(k-1)$, which is easily met by most real--world data. 

For stratified $\mathsf{IkF}$ the requirement is stricter: every subfold must contain at least one sample from every class. A sufficient (though conservative) bound is $n \geq k(k-1)c$, where $c$ is the number of classes. Otherwise, it cannot be guaranteed that each subfold contains at least one instance from each class. 

Class imbalance places and additional limitation. Let $c_{min}$ be the size of the minority class. To ensure that this class appears in every subfold we need $c_{min}\geq k(k-1)$. For instance, for $k=10$ and $c=3$, a minimum of $n=270$ instances is required, and the minority class must contribute at least 90 instances.

These thresholds for $\mathsf{IkF}$ are mild for most real--world datasets. Only in cases of severe imbalance, the value of $k$ should be reduced (e.g., $k=5$) to satisfy the minority--class constraint without discarding samples or oversampling the data.

\begin{figure*}[h]
\centering
\includegraphics[width=1\textwidth]{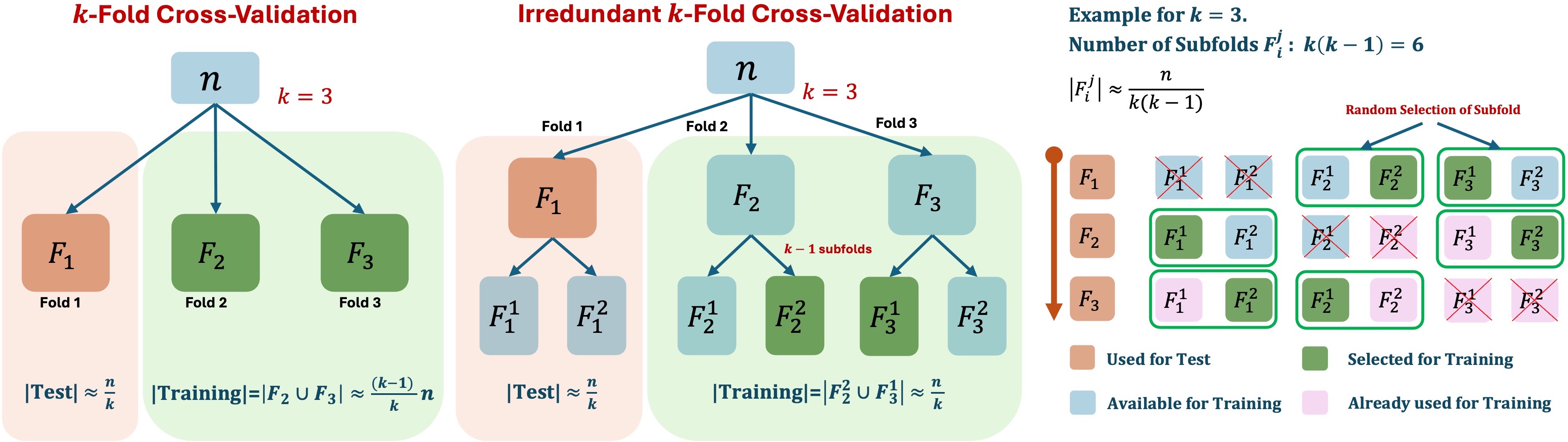}
%\vspace{-0.3cm}
\caption{Graphical comparison between $k$--fold cross--validation and Irredundant $k$--fold cross--validation.}
\label{fig:scheme}
\end{figure*}

Figure~\ref{fig:scheme} illustrates the full $\mathsf{IkF}$ construction for $k=3$. To the left, the traditional $k$--fold cross--validation; in the center, the decomposition for irredundant $k$--fold cross--validation into $k$ folds, which of them is again decomposed into $k\!-\!1$ subfolds. 

To the right, each of the three folds contains two subfolds. In the first iteration, one subfold is selected from each of the other two folds (e.g., $\mathcal{F}_2^2$ and $\mathcal{F}_3^1$), forming the training set for testing on $\mathcal{F}_1$. From fold $\mathcal{F}_2$ we could have selected $\mathcal{F}_2^1$ or $\mathcal{F}_2^2$. Similarly, from fold $\mathcal{F}_3$ we could have selected $\mathcal{F}_3^1$ or $\mathcal{F}_3^2$. In this case, $\mathcal{F}_2^2$ and $\mathcal{F}_3^1$ have been randomly chosen. For test set $\mathcal{F}_2$, the training set will be composed by subfolds from folds $\mathcal{F}_1$ and $\mathcal{F}_3$. From $\mathcal{F}_1$, both subfolds are available, and one of them is randomly selected. However, from $\mathcal{F}_3$ the subfold $\mathcal{F}_3^1$ cannot be chosen because it has already been used (pink) in the previous training set. As subfold reuse is avoided in subsequent iterations, the strategy ensures irredundancy.

\subsection{Computational cost}

A core advantage of $\mathsf{IkF}$ is its lower computational cost. Since each training set is smaller (size $n/k$ vs.\ $(k\!-\!1)n/k$), and each instance is used for training only once, overall computational demand is reduced.

Let $\mathcal{C}(n)$ denote training cost of model $\mathcal{M}$ on $n$ instances. Then, the total training cost over $k$ iterations is
\begin{equation}
\mathcal{C}_{\mathsf{kF}} = k\cdot \mathcal{C}\left((k-1)\frac{n}{k}\right) \qquad \mathcal{C}_{\mathsf{IkF}} = k\cdot \mathcal{C}\left(\frac{n}{k}\right) 
\end{equation}

In traditional $\mathsf{kF}$, each fold uses $n(1 - \frac{1}{k})$ instances for training and $\frac{n}{k}$ for testing. In contrast, $\mathsf{IkF}$ enforces a strict one--to--one usage of instances for training and test across all iterations, resulting in both training and test sets having approximately $\frac{n}{k}$ instances per fold.

%The relative cost reduction depends on the model’s complexity. For linear--time learners (e.g., naïve Bayes), $\mathcal{C}_{\mathsf{kF}} / \mathcal{C}_{\mathsf{IkF}}\!=\!k\!-\!1$; for quadratic models (e.g., support vector machines), the ratio is $(k\!-\!1)^2$. In a typical 10--fold setup, this means up to $9\times$ (linear) or $81\times$ (quadratic) speed--up with respect to the classical $\mathsf{kF}$ approach.

The relative cost reduction depends on the model’s complexity. For linear--time learners (e.g., naïve Bayes), $\mathcal{C}_{\mathsf{kF}} / \mathcal{C}_{\mathsf{IkF}}\!=\!k\!-\!1$. In a typical 10--fold setup, this means up to $9\times$ speed--up with respect to the classical $\mathsf{kF}$ approach.

This makes $\mathsf{IkF}$ particularly appealing in resource--constrained environments, large--scale datasets, or workflows with repeated training (e.g., wrapper--based feature selection or ensemble learning~\cite{Aguilar2024a}). The efficiency gain becomes especially significant for complex models such as generative adversarial networks.

\section{Analysis}\label{sec:analysis}

Let $\mathcal{D}\!=\!\{(x_i,y_i)\}_{i=1}^{n}$ be i.i.d. samples drawn from an unknown joint distribution
$\mathbb{P}_{XY}$, where $x_i\!\in\!\mathcal{X}^d$ and $y_i\!\in\!\mathcal{Y}$. Let $\mathcal{H}$ denote a hypothesis class and $\mathcal{M}$ a deterministic learning algorithm. Given a training set $\mathcal{E}\subset\mathcal{D}$, denote by  $h_\mathcal{E}=\mathcal{M}(\mathcal{E})\in\mathcal{H}$ the model learned by applying $\mathcal{M}$ to $\mathcal{E}$.  

Let $\ell:{\mathcal{Y}}\times{\mathcal{Y}}\to[0,1]$ be $\ell$ a bounded loss function. The true generalization risk of a model $h_\mathcal{E}$ is defined as:
\begin{equation}
\theta = R(h_\mathcal{E})=\mathbb{E}_{(X,Y)\sim\mathbb{P}_{XY}}\bigl[\ell\bigl(Y,h_\mathcal{E}(X)\bigr)\bigr]
\end{equation}
which measures the expected prediction error of $h_\mathcal{E}$ on unseen data.

In a generic $k$--fold cross-validation scheme---either standard or the proposed \emph{irredundant}---the dataset
$\mathcal D$ is partitioned into $k$ mutually–exclusive training--test pairs $(\mathcal E_1,\mathcal T_1),\;(\mathcal E_2,\mathcal T_2),\;\dots,\;(\mathcal E_k,\mathcal T_k)$, where each $\mathcal T_i$ serves as the validation set for the model trained on $\mathcal E_i$.
The fold--wise error estimate and the resulting cross--validation estimator are defined as follows:
\begin{equation}\label{eq:cv-estimates}
\widehat\theta_i
   \;=\;
   \frac{1}{|\mathcal T_i|}
   \sum_{(x,y)\in\mathcal T_i}
   \ell\bigl(y,\,h_{\mathcal E_i}(x)\bigr)
\qquad
\widehat\theta
   \;=\;
   \frac{1}{k}\sum_{i=1}^{k}\widehat\theta_i
\end{equation}

Each $\widehat\theta_i$ estimates the generalization risk $R(h_\mathcal{E})$ for a model trained on a partial dataset and evaluated on an independent test set $\mathcal T_i$. The average $\widehat\theta$ is the standard cross--validation estimator, which aims to approximate the risk of a model trained on the full dataset $\mathcal{D}$, i.e., $R(h_\mathcal{D})$.

In practice, $\widehat{\theta}$ is widely used as a proxy for the generalization error of $h_{\mathcal{D}}$, although this estimator is only unbiased for the expected risk across models trained on subsets of size $|\mathcal{E}_i|$, not necessarily for the risk of the final model trained on $\mathcal{D}$ itself. Nonetheless, under common assumptions—such as model stability or large $n$—$\widehat{\theta}$ typically provides a reliable and consistent estimate in most applications.

Under mild regularity assumptions, both standard and irredundant $k$--fold cross--validation yield consistent estimators of the generalization risk. Let $R(h_n)$ denote the risk of a model trained on $n$ i.i.d. samples. If the learning algorithm $\mathcal{M}$ is \emph{universally consistent}, i.e., $R(h_n) \to R^*$ as $n \to \infty$ for all $\mathbb{P}_{XY}$, and the loss function $\ell$ is bounded, then the cross--validation estimate $\widehat{\theta}$ converges in probability to $R^*$. In particular, for the proposed irredundant procedure $\mathsf{IkF}$, the training set size per fold is $n/k$, which grows with $n$ when $k$ is fixed. Therefore, $\mathsf{IkF}$ remains consistent, since each model in the ensemble is trained on increasingly large samples. Moreover, the absence of training--set overlap in $\mathsf{IkF}$ reduces inter--fold dependence, simplifying variance analysis and enhancing reliability in finite samples. Consequently, $\mathsf{IkF}$ inherits the desirable asymptotic properties of classical cross--validation while providing practical advantages in terms of computational efficiency and variance reduction.

\subsection{Bias Analysis }

The bias of a cross--validation estimator refers to the expected deviation of the estimated performance from the true generalization performance of the model. Formally, if $\theta$ denotes the true expected performance and $\hat{\theta}$ the estimate obtained via cross--validation, the bias is defined as:
\begin{equation}\label{eq:bias}
   \operatorname{Bias}(\hat\theta)=\mathbb E[\hat\theta]-\theta
\end{equation}
where the expectation $\mathbb{E}[\cdot]$ is taken over the randomness induced by the partitioning of the data. This $\operatorname{Bias}(\hat{\theta})\) quantifies the systematic error introduced by evaluating the model on limited training subsets.

In standard $\mathsf{kF}$, the model is trained in each fold on approximately $(1 - \frac{1}{k})n$ instances and evaluated on the remaining $\frac{n}{k}$. Since training sets across folds significantly overlap, most instances contribute to multiple training subsets, which helps reduce the bias: the trained models are effectively exposed to a large portion of the dataset across iterations. As $k$ increases, the size of each training fold approaches $n$, and the resulting models increasingly resemble the one trained on the full dataset. This leads to a small and often negligible bias in the $\mathsf{kF}$ estimator~\cite{Arlot2010}.

In contrast, the proposed $\mathsf{IkF}$ constrains the size of the training set in each iteration to approximately $\frac{n}{k}$, which is significantly smaller than the standard $\mathsf{kF}$ training size. This implies that each model is trained on less information, typically resulting in higher empirical error on unseen data and a potential underestimation of the model’s true generalization performance.  Furthermore, since the training sets are non--overlapping, there is no cumulative benefit from repeated exposure to individual instances across folds.

Therefore, $\mathsf{IkF}$ introduces a slight pessimistic bias---it systematically evaluates models under reduced and independent training conditions. While this results in performance estimates that may be more conservative, such bias is informative in scenarios where robustness and generalization under data scarcity are critical. In particular, it offers a more cautious estimate of performance variability, making it better suited for applications in safety--critical domains or uncertainty quantification.

In summary, standard $\mathsf{kF}$ is nearly unbiased due to overlapping and larger training sets, whereas $\mathsf{IkF}$ may yield slightly more biased estimates due to disjoint, smaller training subsets. However, this conservative bias can be desirable in practice, as it leads to more realistic and robust assessments of model performance under minimal information exposure.

\subsection{Variance Analysis}

One of the key criteria for assessing the reliability of a cross--validation scheme is the variance of the performance estimates it produces. Let $\hat{\theta}_1, \dots, \hat{\theta}_k$ be the performance estimates obtained at each validation iteration. The variance of the cross--validation estimator $\hat{\theta}$ is given by
\begin{equation}\label{eq:variance}
\text{Var}(\hat{\theta}) = \mathbb{E}\left[\left( \hat{\theta} - \mathbb{E}[\hat{\theta}] \right)^2\right]
\end{equation}
where the expectation $\mathbb{E}[\cdot]$ is taken over all possible random partitions of the data into training and testing subsets, according to the cross-validation scheme. The quantity $\text{Var}(\hat{\theta})$ quantifies the sensitivity of the estimate to the specific way the data is split into folds.

High variance indicates that the model's performance estimate is highly sensitive to the chosen data partition, which undermines the robustness of conclusions. In traditional $\mathsf{kF}$ cross--validation, each instance is used for training in $k\!-\!1$ folds and for testing in one fold. Because training sets might be large and largely overlapping, the variance between fold estimates is often low---primarily due to the shared training samples across iterations. However, this overlap reduces the diversity of the training sets and may lead to over--optimistic estimates of generalization.

In general, the use of standard $\mathsf{kF}$ can lead to uneven model behavior, mainly due to \textit{redundant learning}: since each instance appears in the training set of $k\!-\!1$ folds, the same examples are learned repeatedly. This redundancy artificially lowers the observed variance in performance estimates, as the model benefits multiple times from having seen the same samples. As a result, $\mathsf{kF}$ can underestimate the variance that would be observed in real deployment on truly unseen data \cite{Bengio2004}.

As a consequence of the substantial training overlapping in $\mathsf{kF}$, the fold-level estimates $\hat{\theta}_1, \dots, \hat{\theta}_k$ are not statistically independent. The variance of the cross-validation estimator then follows the general decomposition:
\begin{equation}\label{eq:variance_plus}
\text{Var}(\hat{\theta})=\frac{1}{k^2} \left( \sum_{i=1}^{k} \text{Var}(\hat{\theta}_i)+ 2\!\!\!\!\!\!\sum_{1 \leq i < j \leq k}\!\!\!\!\! \text{Cov}(\hat{\theta}_i, \hat{\theta}_j) \right)
\end{equation}
where $\text{Var}(\hat{\theta}_i)$ is the variance of the performance measure on fold $i$, and $\text{Cov}(\hat{\theta}_i, \hat{\theta}_j)$ is the covariance between the fold estimates for $i$ and $j$.

In contrast, the $\mathsf{IkF}$ method ensures that each instance appears in exactly one training set and one test set, thus avoiding any overlap between training and test data across iterations. As a result, training sets are mutually disjoint and smaller in size, which increases diversity but may introduce higher variance---yet more realistic---in individual fold estimates.
may arise from training on almost identical subsets, as in $\mathsf{kF}$.

However, this increased per--fold variance is counterbalanced by the elimination of inter--fold covariance. In standard $\mathsf{kF}$, the proportion of overlap between any two training sets is exactly $(k - 2)/k$; for example, 80\% when $k = 10$. This overlap induces a positive covariance between fold estimates, thereby inflating the overall variance of the estimator. In $\mathsf{IkF}$, the training sets are completely disjoint, and the test sets do not influence other iterations, so the covariances $\text{Cov}(\hat{\theta}_i, \hat{\theta}_j)$ are negligible or zero.

This contrast is crucial: although $\mathsf{IkF}$ may exhibit slightly higher variance at the individual fold level due to reduced training size, its overall estimator variance may be smaller or more representative, thanks to the elimination of cross--fold dependence. In $\mathsf{kF}$, the shared training data leads to positive covariances that bias the variance estimate downward if independence is falsely assumed.

Therefore, the increased variance observed in $\mathsf{IkF}$ should not be seen as a drawback, but rather as a more faithful reflection of the estimator's robustness under realistic, non--redundant training conditions. It offers a more informative evaluation of generalization performance---particularly valuable in high--stakes applications where variability in training data must be fully accounted for.

\subsection{Bias--Variance Trade--off}

The mean squared error (MSE) of the cross--validation estimator \(\hat{\theta}\) with respect to the true generalition error \(\theta\) admits the standard bias--variance decomposition:
\begin{equation}\label{eq:mse_1}
\mathbb{E}\left[\left( \hat{\theta} - \theta \right)^2\right] = \left( \mathbb{E}[\hat{\theta}] - \theta \right)^2 + \mathbb{E}\left[\left( \hat{\theta} - \mathbb{E}[\hat{\theta}] \right)^2\right]
\end{equation}
where the first term, \((\mathbb{E}[\hat{\theta}] - \theta)^2\), corresponds to the squared bias of the estimator, and the second term, \(\text{Var}(\hat{\theta})\), represents its variance. That is:
\begin{equation}\label{eq:mse_2}
\text{MSE}(\hat{\theta}) = \text{Bias}(\hat{\theta})^2 + \text{Var}(\hat{\theta})
\end{equation}
The bias component (Eq. \ref{eq:bias}) arises because in cross--validation, models are trained on subsets of the available data,  resulting in systematically higher errors than those trained on the full dataset. This structural underfitting introduces a negative bias. In contrast, the variance term (Eq. \ref{eq:variance}) captures the sensitivity of the performance estimate  \(\hat{\theta}\) to the random partitioning of the data into folds.

\begin{figure}[tb]
\centering
\includegraphics[width=0.6\linewidth]{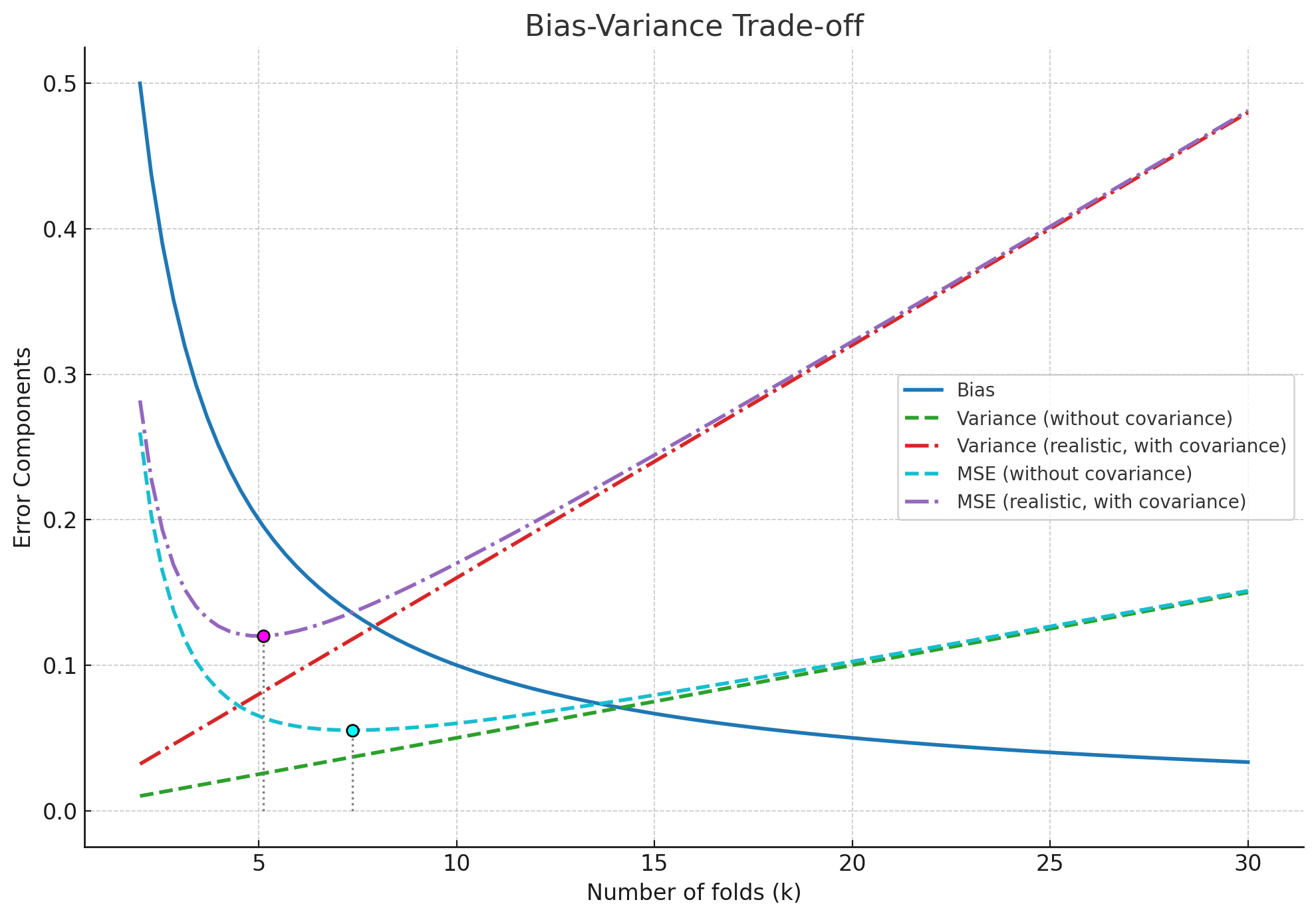}
%\vspace{-0.2cm}
\caption{Variance--Bias Trade--off.}
\label{fig:variance-bias}
\end{figure}

Generally, the variance of $\hat{\theta}$ tends to increase with the number of folds $k$---since each test set becomes smaller and each training set slightly more variable---and decrease as $k$ decreases (i.e., when test sets are larger and folds are more stable). At small $k$ (e.g., $k=5$), bias increases because each model is trained on significantly less data, but variance decreases due to lower instability. At the other extreme, leave-one-out cross-validation ($k=N$) yields minimal bias—since each model is trained on nearly all data---but high variance due to high sensitivity to the specific omitted sample.

Figure~\ref{fig:variance-bias} illustrates this trade-off: the bias (blue solid line) and variance (green dashed line) jointly contribute to the total error (cyan dashed line). However, this classical view assumes fold--wise independence. When inter--fold covariance is correctly accounted for, as in the red dashed line for variance, the true error (purple dashed line) shifts accordingly. Notably, the theoretical minimum MSE (marked by red and cyan circles) differs between the independence and covariance--aware regimes. Neglecting covariance often results in selecting a higher--than--optimal $k$.

This has critical implications for cross--validation strategies. In standard $\mathsf{kF}$, overlapping training sets induce positive covariances between fold estimates, artificially lowering observed variance. In contrast, the novel $\mathsf{IkF}$ scheme eliminates this redundancy by ensuring that each instance is used exactly once for training and once for testing. This structure breaks the dependency between training folds, leading to zero or negligible inter--fold covariance. As a result, the variance of $\hat{\theta}$ becomes more representative of true model instability and less prone to underestimation.

In summary, the bias--variance trade--off fundamentally shapes the reliability of cross--validation estimators. An informed choice of $k$, combined with awareness of model stability and training--set overlap, is essential to minimize mean squared error. The $\mathsf{IkF}$ approach provides a principled solution to mitigate covariance--induced variance distortion while maintaining competitive bias.

\section{Experimental Analysis}\label{sec:experiments}

\begin{table*}[t]
\caption{Performance of the Irredundant $k$--fold CV ($\mathsf{IkF}$) and the standard $k$--fold CV ($\mathsf{kF}$) methods across multiple datasets using the Random Forest classifier. \#s, \#v and \#c stand for the number of samples, variables and classes, respectively. For each method, accuracy (Acc), F--score (Fsc), and running Time (in sec.) are reported. Relative ratios $\mathsf{kF}$/$\mathsf{IkF}$ are shown in the last three columns. Last row shows the mean of each column.}
\label{tab:validation_comparison}
\centering
\fontsize{8}{11}\selectfont
\vspace{0.4cm}
\begin{tabular}{lccc@{\hspace{6mm}}ccc@{\hspace{6mm}}ccc@{\hspace{6mm}}ccc}
\toprule
& & & 
& \multicolumn{3}{c}{\textbf{Irredundant $k$--Fold}}
& \multicolumn{3}{c}{\textbf{Standard $k$--Fold}}
& \multicolumn{3}{c}{\textbf{Ratios \(\mathsf{kF}/\mathsf{IkF}\)}}\\
\textbf{Dataset} & \textbf{\#s} & \textbf{\#v} & \textbf{\#c} 
& \textbf{Acc} & \textbf{Fsc} & \textbf{Time} 
& \textbf{Acc} & \textbf{Fsc} & \textbf{Time} 
& \textbf{RAcc} & \textbf{RFsc} & \textbf{Speed--up} \\
\midrule
magic\_gamma       & 19020 & 10  & 2  & 0.869 & 0.903 & 3.64 & 0.880 & 0.910 & 18.40 & 1.012 & 1.008 & 5.05\\
phishing\_websites & 11055 & 30  & 2  & 0.953 & 0.958 & 0.63 & 0.972 & 0.975 &  1.58 & 1.020 & 1.018 & 2.50\\
room\_occupancy    & 10129 & 16  & 4  & 0.995 & 0.995 & 0.55 & 0.998 & 0.998 &  1.50 & 1.002 & 1.002 & 2.72\\
musk\_v2           &  6598 & 166 & 2  & 0.948 & 0.808 & 1.91 & 0.978 & 0.924 &  7.56 & 1.032 & 1.144 & 3.95\\
landsat\_satellite &  6435 & 36  & 6  & 0.891 & 0.888 & 0.99 & 0.915 & 0.912 &  3.40 & 1.027 & 1.028 & 3.45\\
rice\_cammeo       &  3810 &  7  & 2  & 0.922 & 0.908 & 0.47 & 0.924 & 0.911 &  1.34 & 1.003 & 1.003 & 2.84\\
customer\_churn    &  3150 & 13  & 2  & 0.933 & 0.762 & 0.39 & 0.956 & 0.852 &  0.83 & 1.025 & 1.117 & 2.10\\
cardiotocography   &  2126 & 23  & 10 & 0.829 & 0.821 & 0.42 & 0.891 & 0.889 &  0.92 & 1.075 & 1.083 & 2.17\\
hepatitis\_c       &  1385 & 28  & 4  & 0.246 & 0.245 & 0.41 & 0.261 & 0.260 &  1.04 & 1.059 & 1.060 & 2.55\\
data\_banknote     &  1372 &  4  & 2  & 0.977 & 0.974 & 0.30 & 0.991 & 0.990 &  0.48 & 1.015 & 1.017 & 1.59\\
\midrule
\textbf{Mean}      & \textbf{6508} & \textbf{33} & \textbf{3.6} 
& \textbf{0.856} & \textbf{0.826} & \textbf{0.97}
& \textbf{0.877} & \textbf{0.862} & \textbf{3.71}
& \textbf{1.027} & \textbf{1.048} & \textbf{2.89}\\
\bottomrule 
\end{tabular}
%\vspace{0.13cm}
\end{table*}

We benchmarked \(\mathsf{IkF}\) and \(\mathsf{kF}\) on ten publicly available datasets from the \emph{UCI Machine--Learning Repository} \cite{Dua:2019}. Several properties make these datasets particularly suitable for the present analysis. First, they vary substantially in sample size, from relatively small datasets (e.g., \texttt{hepatitis\_c} with 1,385 samples) to larger datasets (e.g., \texttt{magic\_gamma} with 19,020 samples), allowing the evaluation of validation methods across different sizes. Second, the number of predictive variables covers a wide spectrum, from low--dimensional problems (e.g., \texttt{data\_banknote} with 4 variables) to high--dimensional ones (e.g., \texttt{musk\_v2} with 166 variables), providing a testbed for studying the impact of dimensionality. Third, the classification tasks include both binary and multiclass problems, ensuring that the methods are assessed under varying levels of label complexity. These characteristics collectively ensure that the selected datasets offer a representative and challenging environment for comparing the performance and computational efficiency of the proposed Irredundant $k$--fold validation against the standard $k$--fold approach. A concise overview of these datasets appers in Table \ref{tab:validation_comparison}.

All experiments used 5 folds ($k\!=\!5$), and employed Random Forest classifier with default hyper--parameters implemented in \texttt{scikit-learn} Python package \cite{pedregosa2011scikit}, and were run on an Apple M1 Ultra (20 cores).

Table~\ref{tab:validation_comparison} presents a detailed comparison between the Irredundant $k$--fold (\(\mathsf{IkF}\)) and the standard $k$--fold (\(\mathsf{kF}\)) cross--validation methods across a diverse selection of datasets. For each method, we report the obtained accuracy (Acc), F--score (Fsc), and computational Time (in seconds). Additionally, we provide the relative ratios \(\mathsf{kF}/\mathsf{IkF}\) for each metric, namely RAcc, RFsc, and Speed--up. The last row shows the average values across all datasets.

The results clearly demonstrate that the \(\mathsf{IkF}\) approach achieves performance metrics (accuracy and F--score) that are closely comparable to those obtained by the standard \(\mathsf{kF}\) validation. The mean accuracy ratio \(\text{RAcc}=1.027\) and the mean F--score ratio \(\text{RFsc}=1.048\) are both close to~1, indicating that, on average, removing redundant training instances does not degrade performance.

The performance was also similar when probabilistic measures were used. For instance, for the multiclass classification performance (MCP) curve \cite{Aguilar2022a,Aguilar2024b, Aguilar2024c}, the area under the curve was calculated, providing a mean value of 0.751 for \(\mathsf{IkF}\) and of 0.792 for \(\mathsf{kF}\), which is consistent with the two previous ones. 

A key advantage of \(\mathsf{IkF}\) lies in its significant reduction in computational cost. On average, it reduces the total validation time by a factor of approximately 2.9 compared to \(\mathsf{kF}\), as reflected by the mean Speed--up value. Equivalent gains were also observed (results not shown) with two additional learners---$k$--nearest neighbours and Naïve--Bayes---indicating that the savings are classifier--agnostic. Crucially, this acceleration is obtained without meaningful loss of predictive quality: the average accuracy and F--score differ from those of standard \(\mathsf{kF}\)  by only +2.7\% and +4.8\%, respectively. 

%The consistency of this behavior across datasets with different sizes, numbers of features, and numbers of classes highlights the robustness of the \(\mathsf{IkF}\) method as an efficient alternative to traditional $k$--fold cross--validation, particularly in scenarios where computational resources are limited or when rapid model evaluation is critical.

The consistency of these findings over datasets that vary widely in sample size, dimensionality, and class cardinality underlines the robustness of \(\mathsf{IkF}\) . Consequently, the irredundant method offers a reliable, efficient alternative to classical \(\mathsf{kF}\)---particularly valuable when resources are constrained or rapid model assessment is required.

%%%%%%%%%%%%%%%%%%%%%%%%%%%%%%%%%%%%%%%%%%%%%%%%%%%%%%%%%%%%%%%%%%%%%%%%
\vspace{-0.2cm}

\section{Conclusions}\label{sec:conclusions}

The proposed \emph{Irredundant $k$-fold cross--validation} retains the conceptual simplicity of classical $k$--fold cross--validation while removing a single but consequential source of redundancy: no instance ever appears in more than one training fold. This simple constraint confers several main advantages.

The irredundant structure of $\mathsf{IkF}$ eliminates overlap among training sets, thereby nullifying inter--fold covariance and contributing to reducing the overall variance of the cross--validation estimator. 

While some increase in bias may occur due to reduced training sizes, the nciteet effect is a more stable and interpretable estimate of ge\-ne\-ralization performance, which is preferable in scenarios where stability and low variance are prioritized. This aligns with the pursuit of reproducible and trustworthy model evaluation in machine learning.

Although each model is trained on $n/k$ rather than $(k-1)n/k$ samples---introducing a slightly conservative bias---the empirical gap in accuracy and F--score is marginal ($\approx$2.7 and $\approx$4.8, respectively). In settings where under--fitted behavior is preferable to over--optimism (e.g., medical AI validation), the $\mathsf{IkF}$ validation properties can be an advantage.

The irredundant design of $\mathsf{IkF}$ impacts remarkably on the computational cost of the validation ($\approx$$2.9\times$ speed--up), which might be highly significant for complex models such as those based on neural architectures (e.g., convolutional neural networks, generative adversarial networks, transformers, etc.).

These findings validate the theoretical expectation: by eliminating training--set redundancy, $\mathsf{IkF}$ reduces computational cost while keeping estimator bias and variance in line with those of traditional $\mathsf{kF}$ cross--validation. Consequently, $\mathsf{IkF}$ is a compelling choice whenever limited compute resources or energy consumption are paramount.

Future work will conduct a formal bias--variance analysis of the $\mathsf{IkF}$ under $\beta$--uniform stability \cite{Bousquet2002}, explore deterministic heuristics for assembling the irredundant training sets, and undertake a systematic comparison with alternative resampling schemes across a broad spectrum of dataset sizes, dimensionalities, class cardinalities, and classifier families.

\section*{Acknowledgment}

This work was supported by Grants PID2020-117759GB-I00 and PID2023-152660NB-I00 funded by the Ministry of Science, Innovation and Universities.

\bibliographystyle{unsrt}  
\bibliography{mybibfile}

%\begin{thebibliography}{1}

%\end{thebibliography}

\end{document}